\documentclass{article}

\pdfoutput=1

    \usepackage[preprint, nonatbib]{neurips_2021}

\usepackage[utf8]{inputenc} 
\usepackage[T1]{fontenc}    
\usepackage{hyperref}       
\usepackage{url}            
\usepackage{booktabs}       
\usepackage{amsfonts}       
\usepackage{nicefrac}       
\usepackage{microtype}      
\usepackage{xcolor}         
\usepackage{graphicx,subcaption}       
\usepackage[normalem]{ulem} 
\useunder{\uline}{\ul}{}    
\usepackage{lipsum}
\usepackage{graphicx}

\usepackage{mathtools, amssymb, bm, amsthm}
\usepackage{bbm}
\usepackage{physics}

\usepackage[ruled,vlined]{algorithm2e}

\usepackage{listings}
\usepackage{textcomp}
\usepackage{xcolor}
\usepackage{floatrow}
\newfloatcommand{capbtabbox}{table}[][\FBwidth]
\newcommand{\R}{\mathbb{R}}

\newcommand{\dc}{TiCo}

\newtheorem{theorem}{Remark}[section]

\usepackage{cleveref}

\title{TiCo: Transformation Invariance and \\ Covariance Contrast for Self-Supervised Visual Representation Learning}

\author{
    Jiachen Zhu$^{1}$\thanks{Correspondence to \texttt{jiachen.zhu@nyu.edu}} \And Rafael M. Moraes$^{3}$ \And Serkan Karakulak$^{2}$ \And Vlad Sobol$^{2}$ \And Alfredo Canziani$^{1, 2}$  \And Yann LeCun$^{1,2,4}$ \AND
    \begin{tabular}{l}
        $^1$\normalfont Courant Institute, New York University\\
        $^2$\normalfont Center for Data Science, New York University\\
        $^3$\normalfont Viasat, Inc. \\
        $^4$\normalfont Facebook AI Research
    \end{tabular}
}

\begin{document}

\maketitle

\begin{abstract}
We present \textbf{T}ransformation \textbf{I}nvariance and \textbf{Co}variance Contrast (TiCo) for self-supervised visual representation learning.
Similar to other recent self-supervised learning methods, our method is based on maximizing the agreement among embeddings of different distorted versions of the same image, which pushes the encoder to produce transformation invariant representations.
To avoid the trivial solution where the encoder generates constant vectors, we regularize the covariance matrix of the embeddings from different images by penalizing low rank solutions.
By jointly minimizing the transformation invariance loss and covariance contrast loss, we get an encoder that is able to produce useful representations for downstream tasks.
We analyze our method and show that it can be viewed as a variant of {MoCo} \cite{he2020momentum} with an implicit memory bank of unlimited size at no extra memory cost. This makes our method perform better than alternative methods when using small batch sizes. TiCo can also be seen as a modification of Barlow Twins \cite{zbontar2021barlow}. By connecting the contrastive and redundancy-reduction methods together, TiCo gives us new insights into how joint embedding methods work.

\end{abstract}

\section{Introduction}

The field of self-supervised visual representation learning has seen enormous progress in recent years.
Most successful approaches fall into one of two classes: \emph{pretext task} methods \cite{vincent2008extracting, pathak2016context, zhang2016colorful, zhang2017split, dosovitskiy2014discriminative, doersch2015unsupervised, noroozi2016unsupervised, wang2015unsupervised, pathak2017learning} or \emph{joint embedding} methods \cite{he2020momentum, chen2020simple, misra2020self, caron2020unsupervised, caron2018deep, grill2020bootstrap, chen2020exploring, zbontar2021barlow, bardes2021vicreg}.
Pretext task methods involve training a network to solve a pretext task and using the trained network to generate data representations for downstream tasks that we care about.
A large number of pretext tasks have been proposed, with
some of them reaching state-of-the-art performance at the time of publication.
However, in the past year, multiple joint embedding methods were proposed and their performance surpassed the pretext task methods in almost all standard self-supervised benchmarks.

\begin{figure}[t!]
\begin{subfigure}{1\textwidth} 
    \refstepcounter{subfigure}\label{fig:summary:contrastive}
    \refstepcounter{subfigure}\label{fig:summary:redundancy}
    \refstepcounter{subfigure}\label{fig:summary:clustering}
    \refstepcounter{subfigure}\label{fig:summary:asymmetric}
    \refstepcounter{subfigure}\label{fig:summary:TiCo}
\end{subfigure}%
\centering
\includegraphics[width=\textwidth]{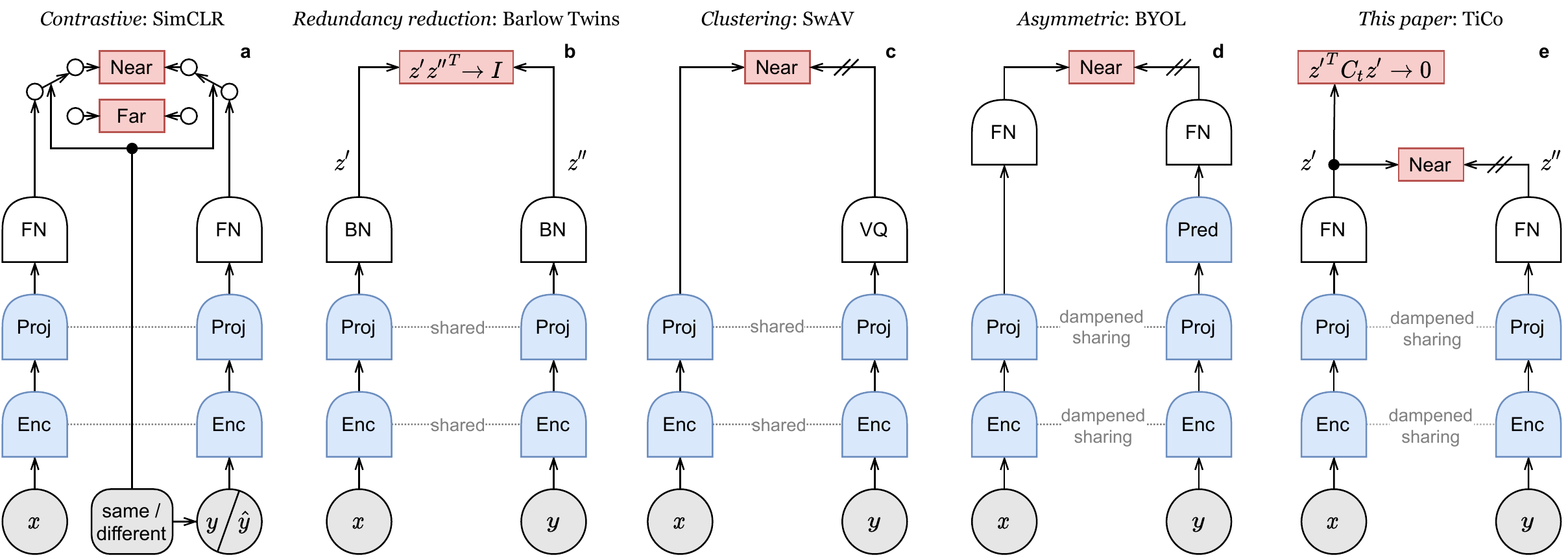}
\caption{
  \textbf{Joint embedding learning categories and this paper's contribution.}
  From left to right, an example of \emph{contrastive learning}, \emph{redundancy reduction}, \emph{clustering-based}, \emph{asymmetric network}, TiCo.
  Shaded circles
  \setlength{\unitlength}{1pt}\begin{picture}(8,8)
  \put(4,3){\textcolor[HTML]{E6E6E6}{\circle*{8}}}\put(4,3){\circle{8}}
  \end{picture}
  represent observed variables, bullet shapes represent deterministic functions, dashed gray lines indicate (dampened) parameter sharing, $\,\mathbin{\!/\mkern-5mu/\!\,}$ indicates a stop-gradient for backpropagation, $\bullet$ represents a bifurcation, and {\fboxsep 1pt\fcolorbox[HTML]{B85450}{F8CECC}{red boxes}} represent cost terms.
  \textbf{Contractions legend:} Enc: encoder, Proj: projector, Pred: predictor, FN: feature normalization, VQ: vector quantization, BN: batch normalization.
}
\label{fig:summary}
\end{figure}

The joint embedding methods rely on the fact that good representations should be invariant to transformations that don't change the semantics of the inputs.
Currently, the majority of joint embedding methods for visual representation learning use a stochastic data augmentation scheme to generate multiple distorted versions of the same image, and push the embeddings of those distorted images towards each other with the objective of making the representations invariant to those data augmentations.
However, simply optimizing such an objective will result in a trivial solution where the encoder generates a constant representation for all inputs, since the constant representation is inherently invariant to any type of transformations in the input space.
Therefore, different methods have been proposed to avoid having such a trivial solution, and they can roughly be summarized into four different categories: \emph{contrastive learning methods} \cite{chen2020simple, he2020momentum, misra2020self}, \emph{clustering-based methods} \cite{caron2018deep, caron2020unsupervised}, \emph{asymmetric network methods} \cite{grill2020bootstrap, chen2020exploring} and \emph{redundancy reduction methods} \cite{zbontar2021barlow, bardes2021vicreg}, as shown in \cref{fig:summary}.

In the present work, we introduce a novel method called TiCo, which also utilizes the joint embedding architecture. We discuss the details of the method in Section \ref{section:method}.
Our method jointly optimizes both the transformation invariance objective and a covariance contrast objective, which effectively regularizes the covariance matrix of the embeddings.
Interestingly, TiCo is both a contrastive learning method and a redundancy reduction method.
On the one hand, it is equivalent to MoCo \cite{he2020momentum} with an altered contrastive loss that enables it to implicitly have an infinitely large memory bank without requiring extra memory, as we show in Section \ref{section:contrastive}.
This allows our method to achieve good performance without an explicit memory bank and use smaller batch sizes than required for other methods.
On the other hand, TiCo can be considered to be the same as the Barlow Twins \cite{zbontar2021barlow} with an exponential moving covariance matrix, which makes it a redundancy reduction method. 
We discuss the similarities in Section \ref{section:redundancy_reduction}.
We demonstrate the effectiveness of TiCo by testing its ability to learn useful representations from ImageNet images in Section \ref{section:results}.
To the best of our knowledge, our work is the first to demonstrate the connection between contrastive learning and redundancy reduction methods. We argue that understanding this connection provides us with a new way to think about the joint embedding learning, which we explain in detail in Section \ref{section:discussion}.

\section{Related Work}

\subsection{Contrastive Learning Methods}

Contrastive learning methods \cite{chopra2005learning, oord2018representation, chen2020simple, he2020momentum} maximize the agreement between representations of different augmentations of the same image while minimizing that of different images, as shown in \cref{fig:summary:contrastive}.
They are able to avoid the trivial solution, since having a constant representation for all images would mean that there is perfect agreement between the representations of different images. The challenge with contrastive learning methods is to find ``hard'' negative pairs that are close enough in the representation space to provide meaningful gradients. This can be addressed by using a large batch size \cite{chen2020simple}, or by sampling negative pairs from representations stored in a memory bank \cite{he2020momentum}. Both of these solutions imply having a large memory footprint. 

\paragraph{Contrastive Loss Function} Many different contrastive loss functions have been proposed \cite{chopra2005learning, oord2018representation, tsai2021self, ozair2019wasserstein, poole2019variational}.
The reason why they are contrastive is because they compare negative and positive pairs of images, where the loss function yields a low value when two positive pairs are close together or two negative pairs are far apart, and a high value vice versa.

\subsection{Redundancy Reduction Methods}

Redundancy reduction methods, like Barlow Twins \cite{zbontar2021barlow}, regularize the covariance matrix of the embeddings to be close to the identity matrix while maximizing the agreement between different augmentations of the same image, as shown in \cref{fig:summary:redundancy}. This formulation avoids having a constant trivial solution, since in that case any pair of components with an equal representation would be perfectly correlated, resulting in having maximum redundancy. 

\subsection{Other Joint Embedding Methods}

Other approaches that recently have shown success in self-supervised representation learning employ clustering-based methods \cite{caron2018deep, caron2020unsupervised} or asymmetric network methods \cite{grill2020bootstrap, chen2020simple}, shown in \cref{fig:summary:asymmetric,fig:summary:clustering}. While clustering based approaches make it easier to mine negative samples and eliminate the need to have a large batch size or a large memory bank, they introduce an additional clustering step that can be computationally expensive. Asymmetric network, like BYOL \cite{grill2020bootstrap}, have shown to have state-of-the-art performance while avoiding producing a constant representation. However, they are hard to analyze and it is not well understood how they avoid the trivial solution.




\section{Method}
\label{section:method}

\subsection{TiCo Algorithm}

\begin{figure}[t!]
\centering
\includegraphics[width=\textwidth]{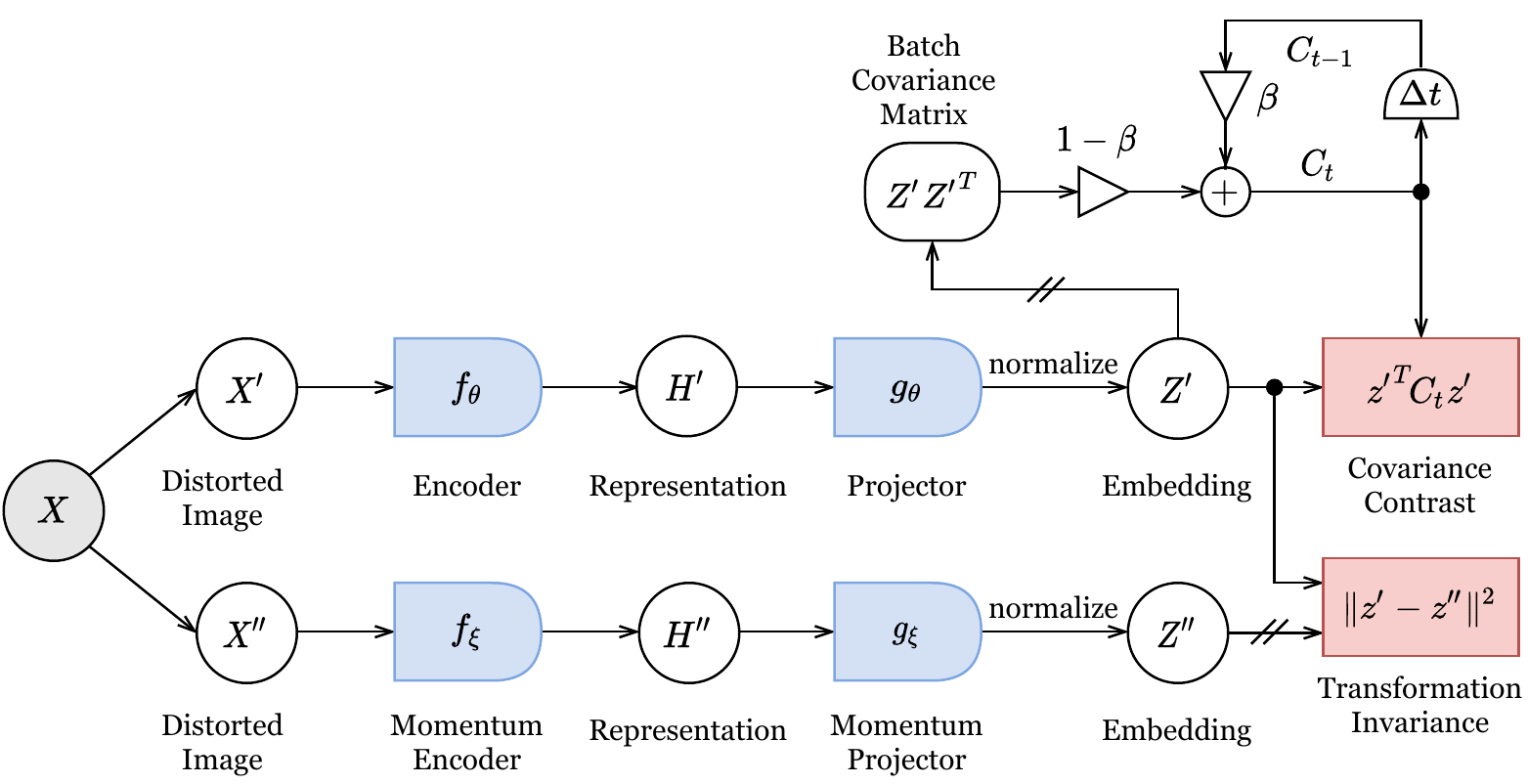}
\caption{
  \textbf{TiCo architecture diagram.}
  Shaded circles
  \setlength{\unitlength}{1pt}\begin{picture}(8,8)
  \put(4,3){\textcolor[HTML]{E6E6E6}{\circle*{8}}}\put(4,3){\circle{8}}
  \end{picture}
  represent observed variables, while empty circles 
  \begin{picture}(8,8)
  \put(4,3){\circle{8}}
  \end{picture}
  represent computed values, bullet shapes represent deterministic functions, $\,\mathbin{\!/\mkern-5mu/\!\,}$ indicates a stop-gradient for backpropagation, $\bullet$ represents a bifurcation, {\fboxsep 1pt\fcolorbox[HTML]{B85450}{F8CECC}{red boxes}} represent cost terms, $\triangle$ represents a multiplication by a scalar value, $\oplus$ is an addition module, and $\Delta t$ represents a unit time delay.
}
\end{figure}

Like other joint embedding methods, TiCo also operates on the embedding pairs of distorted images.
Specifically, given a batch of  $n$ images $X = \{x_1, ..., x_n\}$, two distorted views $X' = \{x'_1, ..., x'_n\}$ and $X'' = \{x''_1, ..., x''_n\}$ are generated using a stochastic data augmentation $\mathcal{T}$.
Then, we use two encoders $f_\theta$ and $f_\xi$ and two projectors $g_\theta$ and $g_\xi$ with parameters $\theta$ and $\xi$ to generate the corresponding embeddings ${z'_i}$ and $ {z''_i}$,  where ${z'_i} = g_\theta(f_\theta(x'_i))$, ${z''_i} = g_\xi( f_\xi(x''_i))$, and ${z'_i}, {z''_i} \in \R^d$.
To simplify the notation, we assume that the outputs of the projector are normalized to unit vectors. 

In our setting, we apply the same momentum encoder technique as proposed in MoCo \cite{he2020momentum}, such that only the parameter $\theta$ is updated through backpropagation, and the parameters $\xi$ is the exponential moving average of the parameter $\theta$. At time step $t$, we have
\begin{align}
\xi_t = \alpha \xi_{t-1} + (1 - \alpha) \theta_t
\end{align}
where $\alpha \in [0, 1]$ is a hyperparameter.

During training, we also keep an exponential moving average of the (non-centered) covariance matrix $C_t$ of the embedding generated by the encoder $f_\theta$ and projector $g_\theta$, such that,
at time step $t$, $C_t$ is updated using the following formula:
\begin{align}
C_t = 
\begin{cases}
\mathbf{0}, & t = 0 \\
\beta C_{t-1} + (1 - \beta) \frac{1}{n} \sum_{i=1}^n {z'_i} {{z'_i}}^T  & \text{otherwise}
\end{cases}
\end{align}
where $\beta \in [0, 1]$ is a hyperparameter.

The final loss function of TiCo has the following form:
\begin{align}
\ell_{\mathrm{TiCo}}(z'_1, ..., z'_n, z''_1, ..., z''_n) &= \frac{1}{2n} \sum_{i=1}^n \|{z'_i} - {z''_i}\|^2 +  \frac{\rho}{n} \sum_{i=1}^{n}  {{z'_i}}^T  C_t {{z'_i}} \\
\label{eq:tico}
&= 1 - \frac{1}{n} \sum_{i=1}^n {z'_i}^T {z''_i} +  \frac{\rho}{n} \sum_{i=1}^{n}  {{z'_i}}^T  C_t {{z'_i}}
\end{align}
where $\rho \in \R^+$ is a hyperparameter that controls the weight of the second term of the loss function.

Intuitively, the loss function jointly optimizes two objectives.
The first term is trying to push the embeddings of different data augmentations of the same image closer to each other.
The second term is trying to push each vector to the subspace of the covariance matrix with smaller eigenvalues.
Therefore, we can decompose the loss function into two parts:
\begin{align}
\ell_{\mathrm{TiCo}}(z'_1, ..., z'_n, z''_1, ..., z''_n) = -   \frac{1}{2n}  \underbrace{\sum_{i=1}^n \|{z'_i} - {z''_i}\|^2}_{\text{transformation invariance}} +  \frac{\rho}{n} \underbrace{ \sum_{i=1}^{n} {{z'_i}}^T  C_t {{z'_i}} }_{\text{covariance contrast}}
\end{align}
which gives us the name of our method: Transformation Invariance and Covariance Contrast. The pseudo code of the TiCo algorithm is shown in \cref{alg:barlow_twins}.

\begin{algorithm}[tb]
   \caption{PyTorch-style pseudocode for TiCo}
   \label{alg:barlow_twins}
   
    \definecolor{codeblue}{rgb}{0.25,0.5,0.5}
    \lstset{
      basicstyle=\fontsize{7.2pt}{7.2pt}\ttfamily\bfseries,
      commentstyle=\fontsize{7.2pt}{7.2pt}\color{codeblue},
      keywordstyle=\fontsize{7.2pt}{7.2pt},
    }
\begin{lstlisting}[language=python, texcl=true, mathescape=true]
# f: encoder network (include both the encoder and projector) with parameter $\theta$
# fm: momentum encoder network (include both the encoder and projector) with parameter $\xi$
# $\rho$, $\alpha$ and $\beta$: hyperparameters
# n: batch size
# d: dimensionality of the embedding
# mm: matrix-matrix multiplication

for x in loader: # load a batch with n samples
    x_1, x_2 = augment(x) # two randomly augmented versions of x
    
    # compute embeddings
    z_1 = f(x_1)  # n$\times$d
    z_2 = fm(x_2) # n$\times$d
    z_1 = F.normalize(z_1, dim=1) # normalize embeddings along the feature dimension
    z_2 = F.normalize(z_2, dim=1) # normalize embeddings along the feature dimension
    
    B = mm(z_1.T, z_1) / n # compute batch covariance matrix
    C = $\beta$C + (1 - $\beta$)B        # update exponential moving covariance matrix
    
    loss = -(z_1 * z_2).sum(dim=1).mean() + $\rho$ * (mm(z_1, C) * z_1).sum(dim=1).mean()

    loss.backward()
    optimizer.step() # optimization step (only $\theta$ is updated)
    
    $\xi$ = $\alpha$$\xi$ + (1 - $\alpha$)$\theta$ # update momentum encoder network (update $\xi$)
    
\end{lstlisting}
\end{algorithm}

\subsection{TiCo as a Contrastive Learning Method}
\label{section:contrastive}

To understand why TiCo is a contrastive learning method, let us first introduce the loss function used by TiCo, which is the squared contrastive loss function below:
\begin{align}\label{eq:sc}
\ell_{\mathrm{Squared}}(z'_1, ..., z'_n, z''_1, ..., z''_n) &= - \frac{1}{n} \sum_{i=1}^n {{z'_i}}^T {{z''_i}} +  \frac{\rho}{n^2} \sum_{i=1}^{n} \sum_{j=1, j \neq i}^{n} \left ({{z'_i}}^T{z''_j} \right )^2 
\end{align}

Let us compare it to the popular InfoNCE contrastive loss function\cite{oord2018representation}: 
\begin{align}
\ell_{\mathrm{InfoNCE}}(z'_1, ..., z'_n, z''_1, ..., z''_n) = - \frac{1}{n} \sum_{i=1}^n {{z'_i}}^T{{z''_i}} + \frac{\tau}{n} \sum_{i=1}^{n} \log \sum_{j=1}^{n} \exp \left ( {{z'_i}}^T{z''_j} / \tau \right )
\end{align}
It is easy to see that, both loss functions pull the positive pairs together in the same way.
However, the way they push negative pairs' embeddings away from each other is different. InfoNCE pushes ${z'_i}$ and $z''_j$ towards the opposite direction of each other to minimize the exponent, while the covariance contrast loss pushes those embeddings in orthogonal directions of each other.
While there is just one direction which is the opposite direction of a given embedding, there are a large number of orthogonal directions in a high dimensional space.
We hypothesize that this makes TiCo loss easier to optimize.

\subsubsection{Combining MoCo with the Squared Contrastive Loss}
Having introduced the squared contrastive loss function, we will now show TiCo's relation to the contrastive learning method MoCo \cite{he2020momentum}.

First, let us substitute the InfoNCE loss used by MoCo with the squared contrastive loss function as below:
\begin{align}
\ell_{\mathrm{Squared}}(z'_1, ..., z'_n, z''_1, ..., z''_m) &= - \frac{1}{n} \sum_{i=1}^n {{z'_i}}^T {{z''_i}} +  \frac{\rho}{nm} \sum_{i=1}^{n} \sum_{j=1}^{m} \left ({{z'_i}}^T{z''_j} \right )^2
\end{align}
where $m$ is the size of the memory bank. We have $m \geq n$ and $\{z''_{n+1}, ..., z''_{m} \}$ are embeddings generated from  previous steps and stored in the memory bank.

It is easy to see that this loss function can be rewritten as follows:
\begin{align}
\ell_{\mathrm{Squared}}(z'_1, ..., z'_n, z''_1, ..., z''_m) &= - \frac{1}{n} \sum_{i=1}^n {{z'_i}}^T {{z''_i}} +  \frac{\rho}{n} \sum_{i=1}^{n}  {{z'_i}}^T \left (\frac{1}{m}\sum_{j=1}^{m} {z''_j} {z''_j}^T \right ) {{z'_i}} \\
&= - \frac{1}{n} \sum_{i=1}^n {{z'_i}}^T {{z''_i}} +  \frac{\rho}{n} \sum_{i=1}^{n}  {{z'_i}}^T  C_t {{z'_i}}
\end{align}
where $C_t = \frac{1}{m}\sum_{j=1}^{m} {z''_j} {z''_j}^T $ at step $t$.

Therefore, instead of saving each vector $z''_j$, which requires memory size $O(md)$, we only need to save the matrix $C_t$ that needs memory size $O(d^2)$.
However, simply replacing the memory bank with the matrix $C_t$ will not work, as it's not obvious how to remove the old embeddings without storing all the $C_t$ matrices from previous $m / n$ steps.
Here, we can apply the same idea of the momentum encoder, and change the matrix $C_t$ from actual average of past $m / n$ batches to an exponential moving average of the outer product $z''_j {z''_j}^T$ as follows:
\begin{align}
C_t = 
\begin{cases}
\mathbf{0}, & t = 0 \\
\beta C_{t-1} + (1 - \beta) \frac{1}{n} \sum_{i=1}^n {z''_i} {z''_i}^[T]  & \text{otherwise}
\end{cases}
\end{align}

After changing the actual average to the exponential moving average, we recover the loss function from equation \ref{eq:tico} used by TiCo.
One difference is that TiCo uses $z' z'^T$ to update the $C_t$, instead of $z'' z''^T$.
That is because,
based on our experiments, using $z' z'^T$ produces a slightly better performance than using $z'' z''^T$.
We leave the analysis of the difference between the two choices as one future work.

The analysis above shows that TiCo is implicitly a contrastive learning method.
Using the momentum encoder and the  exponential moving covariance matrix helps avoiding the necessity of a large batch size or a large memory bank.

\subsection{TiCo as a Redundancy-Reduction Method}
\label{section:redundancy_reduction}

It can be shown that TiCo is also a redundancy-reduction method by demonstrating the similarities between the squared contrastive loss and the loss function used by Barlow Twins. To simplify the discussion, we assume $\theta = \xi$, i.e. that the two encoders are identical.

Let $Z' \in \R^{d \times n}$ be a matrix where $i$-th column is ${z'_i}$, and $Z'' \in \R^{d \times n}$ be a matrix where $i$-th column is ${z''_i}$.
Let ${Z'_F}$ be the matrix $Z'$ normalized along the feature dimension, and ${{Z'_B}}$ be the matrix $Z'$ normalized along the batch dimension.
Then both $\Tr({Z'_F}^T {Z'_F}) = n$ and $\Tr({{Z'_B}} {{Z'_B}}^T) = d$ are constants.

The squared contrastive loss and the loss function used by Barlow Twins are essentially connected through the following results:

\begin{theorem}\label{thm:1}
Given an embedding matrix $Z \in \R^{d \times n}$, the covariance matrix $C = Z Z^T$ and the Gram matrix $K = Z^T Z$ have the following properties:
\begin{enumerate}
\item $C$ and $K$ share the same nonzero eigenvalues $\lambda_1, ..., \lambda_r$, where $r$ is the rank of the $C$ and $K$
\item $\Tr(C) = \Tr(K) = \sum_{i=1}^{\min(n, d)} \lambda_i$
\item $\sum_{i=1}^{d} \sum_{j=1}^d C_{ij}^2 = \|C\|^2_\mathrm{F} = \|K\|^2_\mathrm{F} = \sum_{i=1}^{n} \sum_{j=1}^n K_{ij}^2 = \sum_{i=1}^{\min(n, d)} \lambda_i^2$
\end{enumerate}
\end{theorem}
These are known results of linear algebra, but we give simple proofs in the Appendix.

Since $\theta = \xi$, we can see that $z'_j$ and $z''_j$ have the same distribution.
Then, we can substitute all ${z'_i}^T z''_j$ with ${z'_i}^T z'_j$ without changing the loss function when $i \neq j$ . 

Now, we have:
\begin{align} 
\ell_{\mathrm{Squared}}({Z'_F}, {{Z''_F}}) &= - \frac{1}{n} \sum_{i=1}^n {{z'_i}}^T{{z''_i}} +  \frac{\rho}{n} \sum_{i=1}^{n} \sum_{j=1, j \neq i}^{n} \left ({{z'_i}}^T{z'_j} \right )^2 \label{eqn:1} \\
&= - \frac{1}{n} \Tr({Z'_F}^T {{Z''_F}}) - \rho + \frac{\rho}{n} \|K_F\|^2_\mathrm{F} \\
&= - \frac{1}{n} \Tr({Z'_F}^T {{Z''_F}}) - \rho + \frac{\rho}{n} \|C_F\|^2_\mathrm{F} \label{eq:sctr}
\end{align}
which shows that the loss function is trying to minimize the Frobenius norm of the covariance matrix $\|C_F\|^2_\mathrm{F}$.
Furthermore, we have:
\begin{align}
\|C_F\|^2_F = \sum_{i=1}^{d} \lambda_i^2 \quad \text{ and } \quad \sum_{i=1}^{d} \lambda_i &= n
\end{align}

By Cauchy–Schwarz inequality, we have:
\begin{align}
\left (\sum_{i=1}^{d} \lambda_{i} \right )^2 &\leq \left (\sum_{i=1}^{d} \lambda_{i}^2 \right ) * \left (\sum_{i=1}^{d} 1^2 \right ) \\
\sum_{i=1}^d \lambda_{i}^2 &\geq \frac{n^2}{d}
\end{align}

Hence, the lower bound of $\|C_F\|^2_\mathrm{F}$ is $\frac{n^2}{d}$. It reaches the lowest value when $\lambda_{i} = \frac{n}{d}, \forall i = \{1, ..., d\}$, which means $C_F$ is a scaled identity matrix $\frac{n}{d} I$.
Then, by minimizing the negative part of the squared contrastive loss, we are optimizing the covariance matrix ${Z_F} {Z_F}^T$ to be as close as possible to a scaled identity matrix, in which all the off-diagonal terms are $0$, and all the diagonal terms are equal.
Therefore the redundancy-reduction principle \cite{barlow1961possible} used to justify Barlow Twins \cite{zbontar2021barlow} can also be applied to the loss function of TiCo.

\subsubsection{The Loss Function of Barlow Twins}

Barlow Twins \cite{zbontar2021barlow} has the following loss function:
\begin{align}\label{eq:barlow}
\ell_{\mathrm{Barlow}}({{Z'_B}}, Z''_B) = \frac{1}{d} \sum_{i=1}^d (1 - (C'_B)_{ii})^2 +  \frac{\rho}{d} \sum_{i=1}^{d} \sum_{j=1, j \neq i}^{d} (C'_B)_{ij}^2
\end{align}
where $C'_B = {{Z'_B}} {Z''_B}^T$.

We can do an expansion to the loss function with replacing $C'_B$ with $C_B=  {{Z'_B}} {{{Z'_B}}}^T$ in the second term of equation \ref{eq:barlow}, so we have:
\begin{align}
\ell_{\mathrm{Barlow}}({{Z'_B}}, Z''_B) &= 1 + \frac{1}{d} \sum_{i=1}^d  (C'_B)_{ii}^2 - \frac{2}{d} \sum_{i=1}^d (C'_B)_{ii} + \frac{\rho}{d} \sum_{i=1}^{d} \sum_{j=1, j \neq i}^{d} (C_B)_{ij}^2 \\
&= 1 + \frac{1}{d} \sum_{i=1}^d  (C'_B)_{ii}^2 - \frac{2}{d} \Tr({{{Z'_B}}}^T Z''_B) - \rho + \frac{\rho}{d} \|C_B\|^2_\mathrm{F} \label{eq:bttr}
\end{align}

It is easy to see that there are only two differences between the loss functions in equation \ref{eq:sctr} and equation \ref{eq:bttr}. The first difference is the term $\frac{1}{d} \sum_{i=1}^d  (C'_B)_{ii}^2$.
It can be considered as another regularization term for the covariance matrix, which will not change the optimization objective.
The second difference is the direction of the normalization.
Squared contrastive loss function uses vectors normalized along the feature dimension and the Barlow twins uses values normalized along the batch dimension.
We discuss the effect of the normalization in Section \ref{section:discussion}.

The similarity of TiCo's loss function and the loss function used by Barlow Twins \cite{zbontar2021barlow} further justify the claim that TiCo is a redundancy-reduction method.
Adding the momentum encoder and the exponential moving covariance matrix to our method improves redundancy reduction by decorrelating channels across batches.

\subsection{Implementation Details}

\paragraph{Image augmentations}
We use the same augmentation as used in BYOL \cite{grill2020bootstrap}.
We transform each input image with two sampled augmentations to produce two distorted versions of the input.
The augmentation pipeline consists of random cropping, resizing to $224 \times 224$, randomly flipping the images horizontally,
applying color distortion, optionally converting to grayscale, adding Gaussian blurring,   and applying solarization.

\paragraph{Architecture} The encoder $f_\theta$ is a ResNet-50 model \cite{he2016deep} without final linear layer.
The projector $g_\theta$ consists of
two linear layers, one with $4096$ output units and the other with $256$ output units. We add batch normalization and ReLU between the two linear layers.

\paragraph{Optimization} \label{optim} We closely follow the optimization protocol of BYOL \cite{grill2020bootstrap}. We use the LARS
optimizer \cite{you2017large}. The training schedule starts with a warm-up period which linearly increases the learning rate from 0 to 3.2 ($= 0.2 \times \text{batch size} / 256$) in the first 10 epochs. Then the learning rate slowly decreases to 0.032 ($= 0.002 \times \text{batch size} / 256$) by following cosine decay schedule without restarts. The total number of epochs is $1000$.
Weight decay parameter is set to $1.5 \cdot {10}^{-6}$.
Weight decay and LARS adaptation are not applied to the biases and batch normalization parameters.

\section{Results}
\label{section:results}

\begin{table}[t]
\caption{\textbf{Linear evaluation --- top-1 and top-5 accuracies (in \%), when using a linear classifier on representations. Semi-supervised learning --- top-1 and top-5 accuracies (in \%) when using either 1\% or 10\% of the dataset for training.} All experiments used the validation set of ImageNet. All models used a ResNet-50 architecture as encoder.}
\label{linear_and_semi_table}
\centering


\begin{tabular}{lcccccc}
\toprule
Method & \multicolumn{2}{c}{ Linear Evaluation} & \multicolumn{4}{c}{Semi Supervised} \\
 \cmidrule(lr){2-3}
 \cmidrule(lr){4-7}
 
 & Top-1 & Top-5 & \multicolumn{2}{c}{Top-1} & \multicolumn{2}{c}{Top-5} \\ \cmidrule{4-7}

 &  &  & 1\% & 10\% & 1\% & 10\% \\ 
\midrule

Supervised & $76.5$ &  & $25.4$ & $56.4$ & $48.4$ & $80.4$ \\

\midrule
MoCo & $60.6$ &  &  &  &  &  \\
PIRL & $63.6$ &  &  &  & $57.2$ & $83.8$ \\
SimCLR & $69.3$ & $89.0$ & $48.3$ & $65.6$ & $75.5$ & $87.8$ \\
MoCo v2 & $71.1$ & $90.1$ &  &  &  &  \\
SimSiam & $71.3$ &  &  &  &  &  \\
SwAV & $71.8$ &  &  &  &  &  \\
Barlow Twins & $73.2$ & $91.0$ & $\mathbf{55.0}$ & $69.7$ & $\mathbf{79.2}$ & $89.3$ \\
BYOL & $74.3$ & $\mathbf{91.6}$ & $53.2$ & $68.8$ & $78.4$ & $89.0$ \\
SwAV (with multi-crop)  & $\mathbf{75.3}$ &  & $53.9$ & $\mathbf{70.2}$ & $78.5$ & $\mathbf{89.9}$ \\
TiCo (ours) & $73.4$ & $\mathbf{91.6}$ & $53.0$ & $66.8$ & $\mathbf{79.2}$ & $88.0$ \\
\bottomrule
\end{tabular}%
\end{table}

In order to assess our model's performance, we have used a similar testing regime to Barlow Twins \cite{zbontar2021barlow}.
The training set images of the ImageNet ILSVRC-2012 dataset \cite{deng2009imagenet} are used, without labels, for self-supervised training of our model, following the procedures described in the previous section.
Later the pretrained model evaluated for different tasks, as proposed by \cite{goyal2019scaling}.
In the first experiment, the representations produced by this pretrained encoder are given as input to a linear classifier. In the second experiment, we assess the model's performance with a reduced amount of training data, in a semi-supervised fashion. Experiments 3 and 4 apply the pretrained model to different datasets for classification, object detection and instance segmentation tasks.


\paragraph{ImageNet: Linear Evaluation and Semi-Supervised Learning} For the linear evaluation experiment, we use a ResNet-50 that is pretrained with \dc{} to produce representations of the images in the ImageNet dataset, which are then fed into a linear classifier. For the semi-supervised learning experiments, we sample a subset of images from the ImageNet dataset, either 1\% or 10\% , and use them to fine-tune a ResNet-50 that was pretrained using \dc{}. In \Cref{linear_and_semi_table} we show the top-1 and top-5 accuracies of ours and other state-of-the-art models for each of the experiments.








\textbf{Fixed Representations for Image Classification on Multiple Datasets} We use a model pretrained with \dc{} on ImageNet to produce image representations on a set of other datasets, which are used for training a linear classifier, as suggested by \cite{misra2020self}. This was evaluated on Places-205 \cite{zhou2014learning}, VOC07 \cite{everingham2010pascal} and iNaturalist2018 \cite{van2018inaturalist}. In \Cref{class_table} we show the accuracies (in \%) achieved in each dataset.


\begin{table}[t]
\caption{\textbf{Linear classification using image representations produced by a model pretrained with \dc{}.} Note that the representations are fixed during training. For Places-205 and iNat18, top-1 accuracy (in \%) is reported. For VOC07, the mAP is reported.}
\label{class_table}
\centering
\begin{tabular}{lccc}
\toprule
Method & Places-205 & VOC07 & iNat18 \\
\midrule
Supervised & $53.2$ & $87.5$ & $46.7$ \\
\midrule
SimCLR & $52.5$ & $85.5$ & $37.2$ \\
MoCo v2 & $51.8$ & $86.4$ & $38.6$ \\
SwAV & $52.8$ & $86.4$ & $39.5$  \\
Barlow Twins & $54.1$ & $86.2$ & $46.5$ \\
BYOL & $54.0$ & $86.6$ & $47.6$ \\
SwAV (with multi-crop) & $\mathbf{56.7}$ & $\mathbf{88.9}$ & $\mathbf{48.6}$ \\
TiCo (ours) & $54.0$ & $86.5$ & $45.1$ \\
\bottomrule
\end{tabular}
\end{table}


\paragraph{Object Detection and Instance Segmentation on Multiple Datasets} We have followed the procedures in \cite{he2020momentum} to produce representations for object detection and instance segmentation tasks. In table \Cref{object_table} we present the results on VOC07+12 \cite{everingham2010pascal} and COCO \cite{lin2014microsoft}. In this case, note that the models are fine-tuned during training.

\begin{table}[t]
\caption{\textbf{Object detection and instance segmentation on multiple datasets} 
We apply a model that is pretrained on ImageNet with \dc{} to produce image representations that are used to perform object detection on VOC07+12 and COCO, and instance segmentation on COCO. The object detection task uses Faster R-CNN and the instance segmentation task uses Mask R-CNN, both with the FPN backbone \cite{wu2019detectron2} and the 1x learning rate schedule.
}
\label{object_table}
\centering
\begin{tabular}{lccccccccl}
\toprule
Method & \multicolumn{3}{c}{VOC07+12 det} & \multicolumn{3}{c}{COCO det} & \multicolumn{3}{c}{COCO instance seg} \\
\cmidrule(lr){2-4} \cmidrule(lr){5-7} \cmidrule(lr){8-10}
& AP$_{\textrm{all}}$ & AP$_{50}$ & AP$_{75}$ & AP$^{\textrm{bb}}$ & AP$_{50}^{\textrm{bb}}$ & AP$_{75}^{\textrm{bb}}$ & AP$^{\textrm{mk}}$ & AP$_{50}^{\textrm{mk}}$ & AP$_{75}^{\textrm{mk}}$ \\
\midrule
Supervised & $53.5$ & $81.3$ & $58.8$ & $38.2$ & $58.2$ & $41.2$ & $33.3$ & $54.7$ & $35.2$  \\
\midrule
MoCo v2 & $\mathbf{57.4}$ & $82.5$ & $\mathbf{64.0}$ & $\mathbf{39.3}$ & $58.9$ & $\mathbf{42.5}$ & $34.4$ & $55.8$ & $36.5$  \\
SwAV & $56.1$ & $82.6$ & $62.7$ & $38.4$ & $58.6$ & $41.3$ & $33.8$ & $55.2$ & $35.9$ \\
SimSiam & $57.0$ & $82.4$ & $63.7$ & $39.2$ & $\mathbf{59.3}$ & $42.1$ & $34.4$ & $\mathbf{56.0}$ & $36.7$ \\
Barlow Twins & $56.8$ & $82.6$ & $63.4$ & $39.2$ & $59.0$ & $\mathbf{42.5}$ & $34.3$ & $\mathbf{56.0}$ & $36.5$ \\
TiCo (ours) & $56.2$ & $\mathbf{83.1}$ & $62.3$ & $37.4$ & $57.9$ & $41.0$ & $\mathbf{34.5}$ & $55.2$ & $\mathbf{37.3}$ \\
\bottomrule
\end{tabular}
\end{table}

\section{Discussion}
\label{section:discussion}

\subsection{The Effect of Normalization}

In Section \ref{section:redundancy_reduction}, the normalization can be considered equivalent to adding an equality constraint to the following optimization problem:
\begin{align}
\mathrm{minimize} &\quad \sum_i \lambda_i^2 \\
\mathrm{subject\,to} &\quad \sum_i \lambda_i = 1
\end{align}
Normalizing along the feature dimension or the batch dimension will both result in the same equality constraint, despite having different sets of eigenvalues.
Without this constraint, the optimization will lead to the trivial solution that $\sum_i \lambda_i^2 = 0$.

Recently proposed method VICReg \cite{bardes2021vicreg} doesn't need normalization, since it adds a hinge loss on the standard deviation.
Because the sum of per channel variance is the sum of eigenvalues of the covariance matrix, VICReg can avoid the trivial solution that $\sum_i \lambda_i^2 = 0$ by adding a penalty that prevents the sum of eigenvalues from becoming too small.

\subsection{Gram matrix and Covariance Matrix}

Roughly speaking, we can consider contrastive losses as functions defined on the Gram matrix of the embeddings, and the loss functions used by redundancy reduction methods to be functions defined on the covariance matrix of the embeddings.

The Gram matrix and covariance matrix always share the same eigenvalues.
Therefore, all redundancy reduction methods that only depend on the eigenvalues of the covariance matrix can be thought of contrastive learning methods in general, and contrastive learning methods that only depend on the eigenvalues of the Gram matrix can be considered redundancy reduction methods.

The understanding of the described duality of the Gram matrix and covariance matrix may lead to better designs of loss functions for joint embedding methods.




\section{Conclusion}

In this paper, we have presented TiCo, a self-supervised learning method for visual representation learning.
Despite its simplicity, it achieves similar performance to other state-of-the-art joint embedding methods in multiple tasks and datasets.  We also provide a theoretical explanation of its underlying mechanism.
It can be categorized as both a contrastive learning and a redundancy reduction method.
Lastly, we show that the two types of methods are closely interconnected through the relationship between the Gram matrix and the covariance matrix of the embeddings.

\bibliography{reference}



\appendix

\section{Proof of Remark \ref{thm:1}}

\begin{proof}
\leavevmode
\begin{enumerate}
\item Given an embedding matrix $Z \in \R^{d \times n}$, we can always decompose it using singular value decomposition such that $Z = U \Sigma V^T$, where $U \in \R^{d \times d}, \Sigma \in \R^{d \times n}$, $V \in \R^{n \times n}$, $U$ and $V$ are orthogonal matrices, and $\Sigma$ is a diagonal matrix.

Then we have:
\begin{align}
C &= Z Z^T = U \Sigma V^T V \Sigma^T U^T = U \Sigma \Sigma^T U^T \\
K &= Z^T Z = V \Sigma^T U^T U \Sigma V^T = V \Sigma^T \Sigma V^T
\end{align}

It is easy to see that the columns of $U$ are eigenvectors of $C$, and the columns of $V$ are eigenvectors of $K$ since we have: 
\begin{align}
C u_i = (\Sigma \Sigma^T)_{ii} u_i \\
K v_i = (\Sigma^T \Sigma)_{ii} v_i
\end{align}
where $u_i$ is $i$-th column of $U$ and $v_i$ is $i$-th column of $V$.
The eigenvalues of $C$ and $K$ are the diagonal elements of $\Sigma \Sigma^T$ and $\Sigma^T \Sigma$.

Since $\Sigma$ is a diagonal matrix, $\Sigma \Sigma^T$ and $\Sigma^T \Sigma$ share the same nonzero diagonal elements.
Then $C$ and $K$ share the same nonzero eigenvalues.

\item Since the trace of a matrix is the sum of its eigenvalues, and we already have that $C$ and $K$ share the same nonzero eigenvalues, we have:
\begin{align}
\Tr(C) = \Tr(K) = \sum_i \lambda_i
\end{align}

\item
We have the Frobenius norm of matrices $C$ and $K$ equal to:
\begin{align}
\|C\|_\mathrm{F} &= \sqrt{\sum_i \sum_j C_{ij}^2 } = \sqrt{\sum_i \lambda_i^2} \\
\|K\|_\mathrm{F} &= \sqrt{\sum_i \sum_j K_{ij}^2 } = \sqrt{\sum_i \lambda_i^2}
\end{align}

Since $C$ and $K$ share the same nonzero eigenvalues, it follows that:
\begin{align}
\sum_{i=1}^{d} \sum_{j=1}^d C_{ij}^2 = \|C\|^2_\mathrm{F} = \sum_{i=1} \lambda_i^2 = \|K\|^2_\mathrm{F} = \sum_{i=1}^{n} \sum_{j=1}^n K_{ij}^2 
\end{align}
\end{enumerate}

\end{proof}

\section{Data Augmentations}

For data augmentations, we use the same augmentation parameters as BYOL \cite{grill2020bootstrap} that are listed in \cref{tab:aug}. 

\begin{table}[t]
\caption{\textbf{Data Augmentation Parameters} There are two sets of augmentation parameters. The differences between them are the probabilities of Gaussian blurring and solarization.}
\label{tab:aug}
\centering
\begin{tabular}{lcc}
\toprule
Parameter & $\mathcal{T}$ & $\mathcal{T}'$ \\
\midrule
Random crop probability     & 1.0 & 1.0 \\
Flip probability & 0.5 & 0.5 \\
Color jittering probability     & 0.8 & 0.8 \\
Brightness adjustment max intensity & 0.4 & 0.4 \\
Contrast adjustment max intensity    & 0.4 & 0.4 \\
Saturation adjustment max intensity    & 0.2 & 0.2 \\
Hue adjustment max intensity    & 0.1 & 0.1 \\
Color dropping probability    & 0.2 & 0.2 \\
Gaussian blurring probability    & 1.0 & 0.1 \\
Solarization probability    & 0.0 & 0.2 \\
\bottomrule
\end{tabular}
\end{table}

\paragraph{Removing Augmentations} To examine the effect of removing augmentations, we performed the same ablation studies as proposed in BYOL \cite{grill2020bootstrap}, where all the experiments were done with a batch size of 4096 and for 300 epochs.
The experiments show that our method is robust to changes in data augmentations.
Without using any data augmentation except random cropping, our model only suffered a $11.3\%$ accuracy drop, which is a significant improvement over SimCLR \cite{chen2020simple} that suffers a $27.7\%$ accuracy drop using the same setting.
The result is listed in \cref{tab:test}.

\begin{table}[t]

\centering
\begin{tabular}{lccc}
\toprule
Transformation & BYOL & SimCLR & TiCo \\
\midrule
Baseline     & 72.5 & 67.9 & 71.4 \\
No grayscale & 70.3 & 61.9 & 68.0 \\
No color     & 63.4 & 45.7 & 62.7 \\
Color + blur & 61.1 & 41.7 & 62.6 \\
Crop only    & 60.1 & 40.2 & 60.1 \\
\bottomrule
\end{tabular}
{\caption{\textbf{Top-1 accuracies (in \%) with different data augmentations} Data for SimCLR and BYOL are from \cite{grill2020bootstrap}. All the experiments include random flipping data augmentation.}\label{tab:test}}
\end{table}

\begin{figure}[t!]
\centering
\caption{\textbf{Impact of progressively removing data augmentations} We show the difference of the top-1 accuracy between using the baseline augmentation and using the simpler augmentation with SimCLR, BYOL and TiCo}\label{fig:test}
\includegraphics[width=0.6\textwidth]{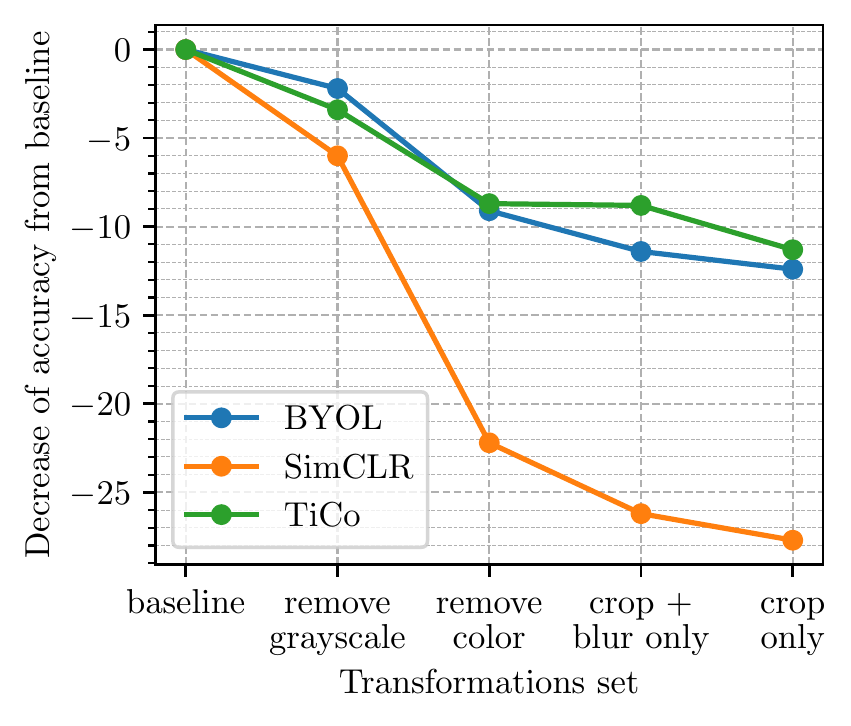}
\end{figure}

  


\section{Hyperparameters}

\subsection{Pretraining}

We have trained the final model for $1000$ epochs using a batch size of $4096$.
For the momentum encoder's hyperparameter $\alpha$, we increased it from $0.99$ to $1.0$ using a cosine decay schedule without restarts.
During the whole training, the covariance matrix hyperparameter $\beta$ is fixed to $0.9$ and the weight hyperparameter $\rho$ for the covariance contrast loss is fixed to $8.0$.
All other hyperparameters are the same as stated in \cref{optim}.

We also apply the weight standardization trick \cite{qiao2019micro} for all convolution layers to accelerate the convergence of training for experiments with fewer epochs.
For 1000 epoch experiments, we don't see a significant improvement over the final accuracy (top-1 accuracy $73.4\%$ with weight standardization and $73.2\%$ without weight standardization).

\subsection{Linear Evaluation on ImageNet}

We follow the same linear evaluation protocol as in BYOL \cite{grill2020bootstrap}.
We fix the weights of the encoder network.
For the linear classifier, we optimize the cross-entropy loss using SGD with Nesterov momentum over $80$ epochs using a batch size of $1024$, with learning rate $0.4$ and momentum $0.9$.
We do not use weight decay.

\subsection{Semi-Supervised Learning on ImageNet}

We follow the same semi-supervised learning protocol as in Barlow Twins \cite{zbontar2021barlow}.
We optimize the cross-entropy loss using SGD with Nesterov momentum over $20$ epochs using a batch size of $256$, with learning rate $0.002$ for the encoder network, learning rate $0.5$ for the classifier network and momentum $0.9$ for both networks.
Both learning rates are multiplied by a factor of $0.2$ after the 12th and 16th epoch.
We do not use weight decay.

\subsection{Linear Evaluation on Other Datasets}

We follow the same transfer learning evaluation protocol as in PIRL \cite{misra2020self}.
We fix the weights of the encoder network for all tasks.
For Places-205
and iNaturalist2018 we train a linear classifier with SGD
($14$ epochs on Places-205, $84$ epochs on iNaturalist2018)
with a learning rate of $0.05$ for Places-205 and $4.0$ for iNaturalist2018, using an SGD momentum of 0.9 for both.
The learning rate of Places-205 is multiplied by a factor of $0.5$ after the 4th, 8th and 12th epochs.
The learning rate of iNaturalist2018 is multiplied by a factor of $0.1$ after the 24th, 48th and 72th epochs.
We do not use weight decay for either of the two tasks.

For VOC07 dataset, we train SVM classifiers where the $C$ values are computed using cross-validation.

\subsection{Object Detection and Instance Segmentation}

We use the VISSL \cite{goyal2021vissl} and detectron2 \cite{wu2019detectron2} libraries for training and evaluating the detection models.

\paragraph{VOC07+12} We use the VOC07+12 \textit{trainval} set for training a Faster R-CNN \cite{ren2016faster} with FPN backnone for $24000$ iterations using a batch size of $16$.
The initial learning rate for the model is $0.15$ which is reduced by a factor of $0.1$ after $18000$ and $22000$ iterations.
We also use linear warmup \cite{goyal2017accurate} for the first 1000 iterations.

\paragraph{COCO} We use the COCO 2017 \textit{train} split to train Mask R-CNN \cite{he2017mask} with FPN backbone. We use a learning rate of $0.08$ and keep
the other parameters the same as in the 1× schedule in detectron2.

\section{Miscellaneous}

\subsection{Compute Resources}

The majority of our experiments were run using AMD MI50 GPUs.
The final pretraining for 1000 epochs takes about 108 hours on 8 nodes, where each node has 8 MI50 GPUs attached.
We estimate that the total amount of compute resources used for all the experiments can be roughly approximated by $75 \text{ (days)} \times 24 \text{ (hours per day)} \times 8 \text{ (nodes)} \times 8 \text{ (GPUs per nodes)} = 115,200 \text{ (GPU hours)}$.

We are aware of potential environmental impact of consuming a lot of compute resources needed for this work, such as atmospheric $\text{CO}_2$ emissions due to the electricity used by the servers. However, we also believe that advancements in self-supervised learning and representation learning can potentially help mitigate these effects by reducing the need for data and compute resources in the future.

\subsection{Limitation of Experiment Results} 

Due to a lack of compute resources, we were not able to conduct a large number of experiments with the goal of tuning hyperparameters and searching for the best configurations.
Therefore, the majority of hyperparameters and network configurations used in this work are the same as provided by BYOL \cite{grill2020bootstrap} or Barlow Twins \cite{zbontar2021barlow}.
The only hyperparameters that were more carefully tuned were $\rho$, the weight of the covariance contrast loss and $\beta$, which controls the update of the covariance matrix.
All the other hyperparameters may not be optimal. 

In addition, all models were pretrained on the ImageNet \cite{deng2009imagenet} dataset, so their performances might differ if pretrained with other datasets containing different data distributions or different types of images (e.g., x-rays). 
We encourage further exploration in this direction for current and future self-supervised learning frameworks.


\end{document}